\documentclass[10pt,twocolumn,letterpaper]{article}

\usepackage[pagenumbers]{cvpr} 

\usepackage{amssymb}
\usepackage{booktabs}

\usepackage[utf8]{inputenc} 
\usepackage[T1]{fontenc}    
\usepackage{url}            
\usepackage{booktabs}       
\usepackage{amsfonts}       
\usepackage{nicefrac}       
\usepackage{microtype}      
\usepackage{xcolor}         
\usepackage{bm}         
\usepackage{amsmath}         
\usepackage{graphicx}

\usepackage{multirow}
\usepackage{makecell}
\usepackage{pifont}
\usepackage{wrapfig}

\usepackage{caption}
\usepackage{subcaption}
\usepackage{graphicx}
\usepackage{comment}
\usepackage{xr}

\newcommand{\tabincell}[2]{\begin{tabular}{@{}#1@{}}#2\end{tabular}}
\newcommand{\xmark}{\ding{55}}
\newcommand{\cmark}{\ding{51}}
\newtheorem{proposition}{Proposition}
\newtheorem{definition}{Definition}[section]

\usepackage[pagebackref,breaklinks,colorlinks]{hyperref}

\def\cvprPaperID{***} 
\def\confName{CVPR}
\def\confYear{2023}

\begin{document}

\title{ConsistentNeRF: Enhancing Neural Radiance Fields with 3D Consistency for Sparse View Synthesis}



\author{
Shoukang Hu\textsuperscript{\rm 1} \quad
Kaichen Zhou\textsuperscript{\rm 2} \quad
Kaiyu Li\textsuperscript{\rm 1} \quad
Longhui Yu\textsuperscript{\rm 3} \quad
Lanqing Hong\textsuperscript{\rm 4} \quad
Tianyang Hu\textsuperscript{\rm 4} \\
Zhenguo	Li\textsuperscript{\rm 4}\footnotemark[1] \quad
Gim Hee	Lee\textsuperscript{\rm 5}\footnotemark[1] \quad
Ziwei Liu\textsuperscript{\rm 1}\footnotemark[1] \\
\textsuperscript{\rm 1} Nanyang Technological University \quad
\textsuperscript{\rm 2} University of Oxford \quad
\textsuperscript{\rm 3} Peking University \\
\textsuperscript{\rm 4} Huawei Noah’s Ark Lab \quad
\textsuperscript{\rm 5} National University of Singapore
}

\maketitle

\renewcommand{\thefootnote}{\fnsymbol{footnote}}
\footnotetext[1]{Joint last authorship.}

\begin{abstract}
Neural Radiance Fields (NeRF) has demonstrated remarkable 3D reconstruction capabilities with dense view images. However, its performance significantly deteriorates under sparse view settings. We observe that learning the 3D consistency of pixels among different views is crucial for improving reconstruction quality in such cases. In this paper, we propose ConsistentNeRF, a method that leverages depth information to regularize both multi-view and single-view 3D consistency among pixels. 
Specifically, ConsistentNeRF employs depth-derived geometry information and a depth-invariant loss to concentrate on pixels that exhibit 3D correspondence and maintain consistent depth relationships.
Extensive experiments on recent representative works reveal that our approach can considerably enhance model performance in sparse view conditions, achieving improvements of up to 94\% in PSNR, 76\% in SSIM, and 31\% in LPIPS compared to the vanilla baselines across various benchmarks, including DTU, NeRF Synthetic, and LLFF.
\end{abstract}

\section{Introduction}
\label{sec:intro}
Novel view synthesis is a longstanding challenge in the fields of computer vision and graphics. The objective is to generate photorealistic images from perspectives that were not originally captured~\cite{jang2021codenerf, li2021mine,liu2021neural,rematas2021sharf}. Recently, the employment of coordinate-based representation learning in 3D vision has increased the popularity of novel view synthesis. Neural Radiance Fields (NeRF)\cite{mildenhall2020nerf} serves as a notable example that leverages a coordinate-based neural network and dense proximal views to yield high-quality and realistic outcomes. However, NeRF's capacity for realistic novel view synthesis is constrained in sparse view settings, due to the insufficiency of supervisory information and the inherent challenges of learning 3D consistency from limited data~\cite{huang2022stylizednerf}. This limitation leads to unsatisfactory performance and restricts the method's applicability in real-world situations.
To address the limitations of NeRF in sparse view settings, researchers have proposed two main strategies. The first strategy involves pre-training NeRF on large-scale datasets containing multiple scenes and subsequently fine-tuning the model~\cite{chen2021mvsnerf,chibane2021stereo, jang2021codenerf, li2021mine,liu2021neural,rematas2021sharf,trevithick2021grf, wang2021ibrnet, yu2021pixelnerf}. The second strategy introduces additional regularization to optimize NeRF~\cite{deng2021depth, jain2021putting, roessle2021dense, niemeyer2021regnerf, wei2021nerfingmvs, kim2022infonerf, wang2023sparsenerf}. However, these approaches tend to focus primarily on pixel-level color and depth within a single view, rather than emphasizing both multi-view and single-view 3D consistency. In contrast, existing works dedicated to other 3D tasks, such as depth estimation and scene synthesis, demonstrate that 3D consistency is particularly important for accurate 3D appearance and geometry reconstruction~\cite{rockwell2021pixelsynth,godard2019digging, zhou2022devnet}.

\begin{figure}[t]
	\centering
    \includegraphics[width=7.5cm]{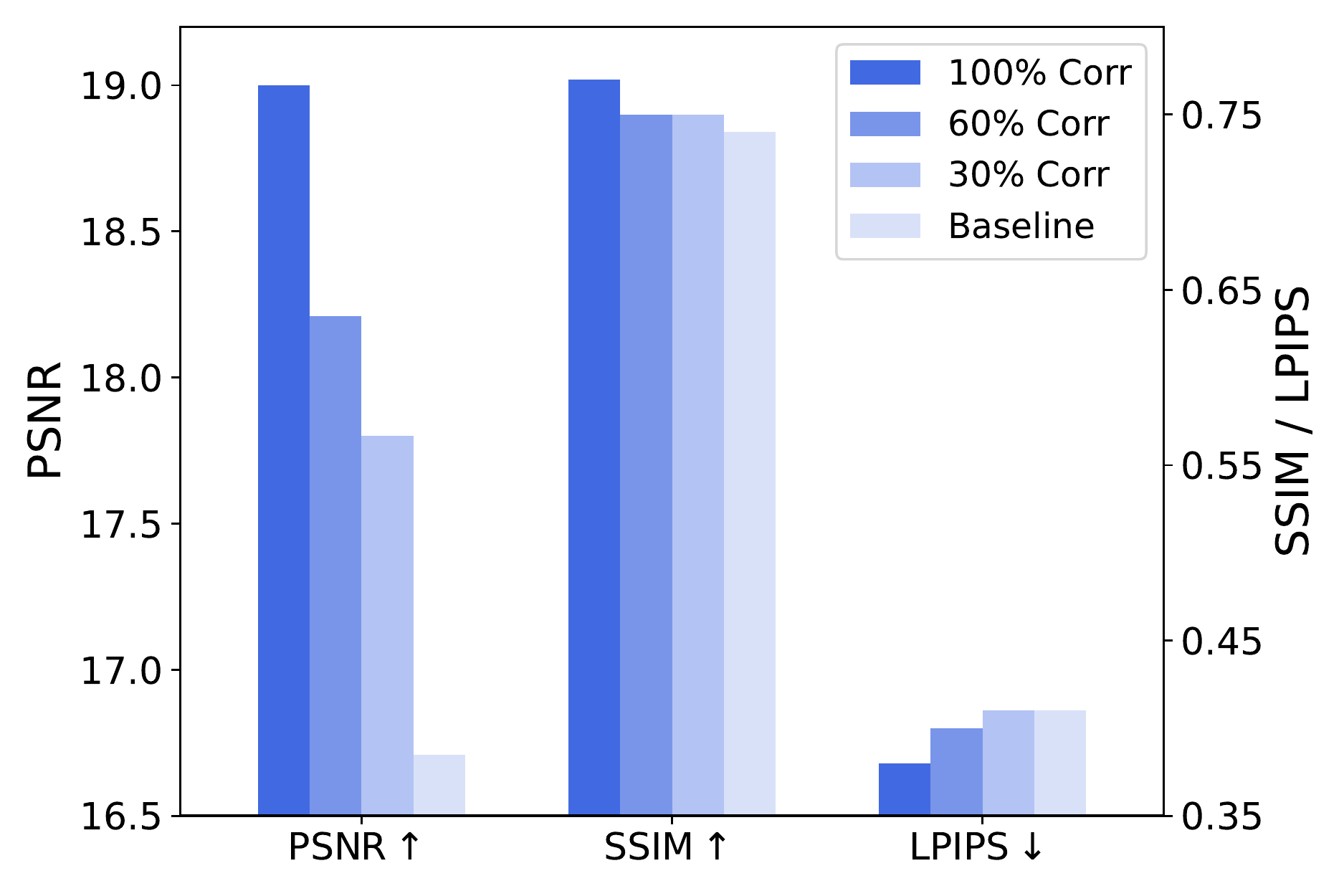}
    \caption{Performance (PSNR$\uparrow$, SSIM$\uparrow$, LPIPS$\downarrow$) comparison of NeRF with different levels of multi-view 3D consistency information. Using more multi-view 3D consistency constraints leads to better model performance.}
	\label{fig: corr_bar}
\end{figure}

\begin{figure*}[t!]
    \centering
    \includegraphics[width=14cm]{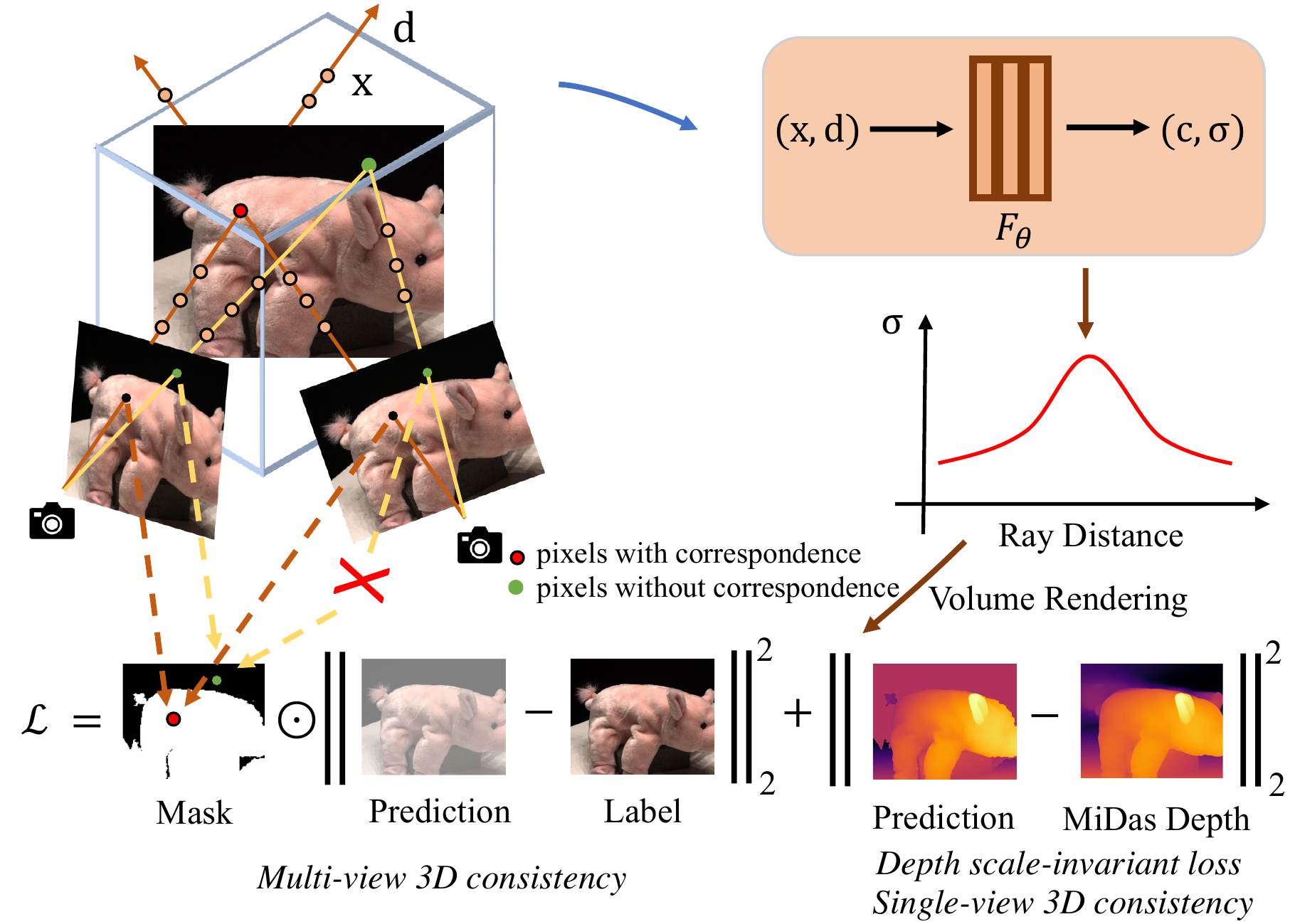}
    \caption{\textbf{The demonstration of proposed multi-view and single-view 3D consistency regularization.} We regularize multi-view 3D consistency by utilizing the multi-view depth correspondence among different views to mask pixels satisfying 3D correspondence (the red point) or not (the green point) and construct the loss based on the mask information. We also regularize single-view 3D consistency by constructing a depth scale-invariant loss function based on the monocular depth predicted from state-of-the-art MiDas model.
    } 
    \label{fig:ConsistentNeRF_overview}
\end{figure*}

In the field of 3D reconstruction, there are two types of 3D consistency relationships: multi-view and single-view 3D consistency. Multi-view 3D consistency refers to the correspondence between pixels that result from projecting the same 3D scene point into different views. To achieve this correspondence, the predicted color and depth must match and satisfy the homography warping relationship, as shown in Fig.\ref{fig: corr_bar}. Our evaluation demonstrates that including increasing amounts of 3D correspondence information into NeRF optimization improves performance in sparse view settings, highlighting the importance of 3D consistency as discussed in Sec.\ref{sec: pre_corr}. Single-view 3D consistency refers to the 3D geometric relationship of pixels within the same view. However, incorporating both multi-view and single-view 3D consistency into NeRF optimization poses a challenging problem. 

In this study, we introduce Consistent Neural Radiance Fields (ConsistentNeRF), a solution that explicitly integrates multi-view and single-view 3D consistency to improve performance in sparse view scenarios. Specifically, to direct NeRF optimization towards pixels that fulfill the multi-view correspondence relationship, ConsistentNeRF selects these pixels based on depth-derived geometric information and assigns higher loss weights during training. For single-view consistency, we utilize a depth-invariant loss to extract 3D consistency information from nearby views employing the DPT Large pre-trained model~\cite{ranftl2020towards}. Our proposed method achieves state-of-the-art results compared to existing approaches, including NeRF~\cite{mildenhall2020nerf}, DSNeRF~\cite{deng2021depth}, Mip-NeRF~\cite{barron2021mip}, InfoNeRF~\cite{kim2022infonerf}, DietNeRF~\cite{jain2021putting}, RegNeRF~\cite{niemeyer2021regnerf}, MVSNeRF~\cite{chen2021mvsnerf}, GeoNeRF~\cite{johari2022geonerf}, and ENeRF~\cite{lin2022efficient}, across various datasets such as DTU dataset~\cite{jensen2014large}, Forward-Facing LLFF dataset~\cite{mildenhall2019local} and Realistic Synthetic NeRF dataset~\cite{mildenhall2020nerf}.

The main contributions of this work contain three parts:
\begin{enumerate}
    \item We introduce ConsistentNeRF, a method that effectively combines multi-view and single-view 3D consistency to improve sparse view synthesis performance.
    \item Our approach utilizes depth-derived geometric information and a depth-invariant loss, achieving state-of-the-art results compared to existing methods across various datasets.
    \item The significant improvements demonstrated by ConsistentNeRF showcase the effectiveness of the proposed method for enhancing 3D consistency in Neural Radiance Fields.
\end{enumerate}

\section{Related Works}
Two methods were proposed to enhance the generalizability of NeRF with sparse-views: incorporating prior knowledge and introducing additional ground-truth information.

\noindent \textbf{View Synthesis with Prior Knowledge.} 
Pre-training neural networks with a large amount of data is a popular approach to incorporate prior knowledge and reduce the need for dense views in rendering novel 3D scenarios. Algorithms such as SSRF~\cite{chibane2021stereo}, GRF~\cite{trevithick2021grf}, Point-NeRF~\cite{xu2022point}, IBRNET~\cite{wang2021ibrnet}, PixelNeRF~\cite{yu2021pixelnerf}, Neural rays~\cite{liu2022neural} and MVSNeRF~\cite{chen2021mvsnerf} use pre-trained models to extract feature maps from source views, which are then used to form appearance and geometry features for points in target views. Despite their effectiveness in dealing with sparse views, these algorithms still experience a significant decrease in performance when tested on scenarios with dense views or sparse views. ConsistentNeRF, on the other hand, improves the performance of models under sparse view settings without adding to the computational burden by incorporating 3D consistency relationships to regulate the optimization process.

\noindent\textbf{View Synthesis with Additional Information.} 
This research introduces additional information to assist the view synthesis process in sparse-view scenarios. DSNeRF~\cite{deng2021depth}, GeoNeRF~\cite{johari2022geonerf} and ENeRF~\cite{lin2022efficient} incorporate geometry constraints using ground truth or "free" depth information. CodeNeRF~\cite{jang2021codenerf}, DoubleField~\cite{shao2022doublefield}, ShaRF~\cite{rematas2021sharf}, Improving~\cite{darmon2022improving}, and DietNeRF~\cite{jain2021putting} introduce object-centric shape or semantic information to build better correspondences among views. RegNeRF ~\cite{niemeyer2021regnerf} and RapNeRF~\cite{zhang2022ray} introduce regularization, but none have used cross-view 3D consistency. In this work, the optimization of NeRF is regularized through 3D consistency relationships. 
Our concurrent work SPARF~\cite{truong2022sparf} applies the network mapping to derive the correspondence relationship among different views.

\section{Method}

\subsection{Background}
\textbf{Neural Radiance Fields.} The Radiance Field learns a continuous function which takes as input the 3D location $\mathbf{x}$ and unit direction $\mathbf{d}$ of each point and predicts the volume density $\mathbf{\sigma} \in [0, \infty)$ and color value $\mathbf{c}\in [0,1]^3$. In NeRF~\cite{mildenhall2020nerf}, this continuous function is parameterized by a multi-layer perception (MLP) network $F_{\theta}: (\gamma(\mathbf{x}), \gamma(\mathbf{d})) \to (\mathbf{c}, \mathbf{\sigma})$, where the weight parameters $\theta$ are optimized to generate the volume density $\mathbf{\sigma}$ and directional emitted color $\mathbf{c}$, $\gamma$ is the predefined positional embedding applied to $\mathbf{x}$ and $\mathbf{d}$, which maps the inputs to a higher dimensional space.

\textbf{Volume Rendering.} 
Given the Neural Radiance Field (NeRF), the color of any pixel is rendered with principles from classical volume rendering~\cite{kajiya1984ray} the ray $\mathbf{r}(t) = \mathbf{o} + t\mathbf{d}$ cast from the camera origin $\mathbf{o}$ through the pixel along the unit direction $\mathbf{d}$.
In volume rendering, the volume density $\sigma(\mathbf{x})$ can be interpreted as the probability density at an infinitesimal distance at location $\mathbf{x}$. With the near and far bounds $t_n$ and $t_f$, the expected color $\hat{C}_{\theta}(\mathbf{r})$ of camera ray $\mathbf{r}(t) = \mathbf{o} + t\mathbf{d}$ is defined as
\begin{equation}
\label{eq:nerf_volume_redndering}
\begin{aligned}
\hat{C}_{\theta}(\mathbf{r})=\int_{t_n}^{t_f} T(t) \sigma(\mathbf{r}(t)) \mathbf{c}(\mathbf{r}(t), \mathbf{d})dt, \\
\text{ where } T(t)=\exp(-\int_{t_n}^{t}\sigma(\mathbf{r}(s))ds),
\end{aligned}
\end{equation}
where $T(t)$ denotes the accumulated transmittance along the direction $\mathbf{d}$ from $t_n$ to $t$. In practice, the continuous integral is approximated by using the quadrature rule~\cite{max1995optical} and reduced to the traditional alpha compositing. The neural radiance field is then optimized by constructing the photometric loss $\mathcal{L}$ between the rendered pixel color $\hat{C}_{\theta}(\mathbf{r})$ and ground truth color $C(\mathbf{r})$:
\begin{equation}
\label{eq:nerf_l2_loss}
\begin{aligned}
\mathcal{L} = \frac{1}{|\mathcal{R}|}\sum_{\mathbf{r}\in \mathcal{R}} ||\hat{C}_{\theta}(\mathbf{r}) - C(\mathbf{r})||_2^2 ,
\end{aligned}
\end{equation}
where $\mathcal{R}$ denotes the set of rays, and $|\mathcal{R}|$ is the number of rays in $\mathcal{R}$.

\subsection{Preliminary: Multi-view Pixel-wise 3D Consistency}
\label{sec: pre_corr}

In this section, we demonstrate the importance of considering the correspondence, i.e., multi-view 3D consistency, in the optimization process.
With no loss of generality, we define $\mathcal{M}$ to be the set containing pixels satisfying correspondence relationship and $\mathcal{T}$ to be the correspondence relationship between pixel $(i,j)$ and $(m,n):=\mathcal{T}((i,j))$.
The 3D multi-view appearance consistency is defined in Definition~\ref{def: appearance}. 
Similarly, we also define the 3D multi-view geometry consistency and details are shown in Appendix.~\ref{app: geometry_consistency_pf}.
By involving the proposed mask in Sec.~\ref{subsec: multi-view 3D concsitency nerf}, we select and assign larger loss weights to pixels that satisfy the homography warping relationship between source views and target views, i.e., the correspondence relationship. 
We compare the performance (PSNR$\uparrow$, SSIM$\uparrow$, LPIPS$\downarrow$) of assigning larger weights to different portions (30\%, 60\%, 100\%) of pixels satisfying the correspondence relationship in the DTU data set.
The baseline is the original NeRF model that treats all pixels equally during the optimization process. 
As shown in Fig.~\ref{fig: corr_bar}, assigning large weights to more pixels satisfying correspondence leads to better model performance.
More details can be found in Appendix~\ref{app:preliminary}.

\begin{figure*}[t]
    \setlength{\abovecaptionskip}{0.2cm}
    \setlength{\belowcaptionskip}{-0.1cm}
    \centering
    \includegraphics[width=14cm]{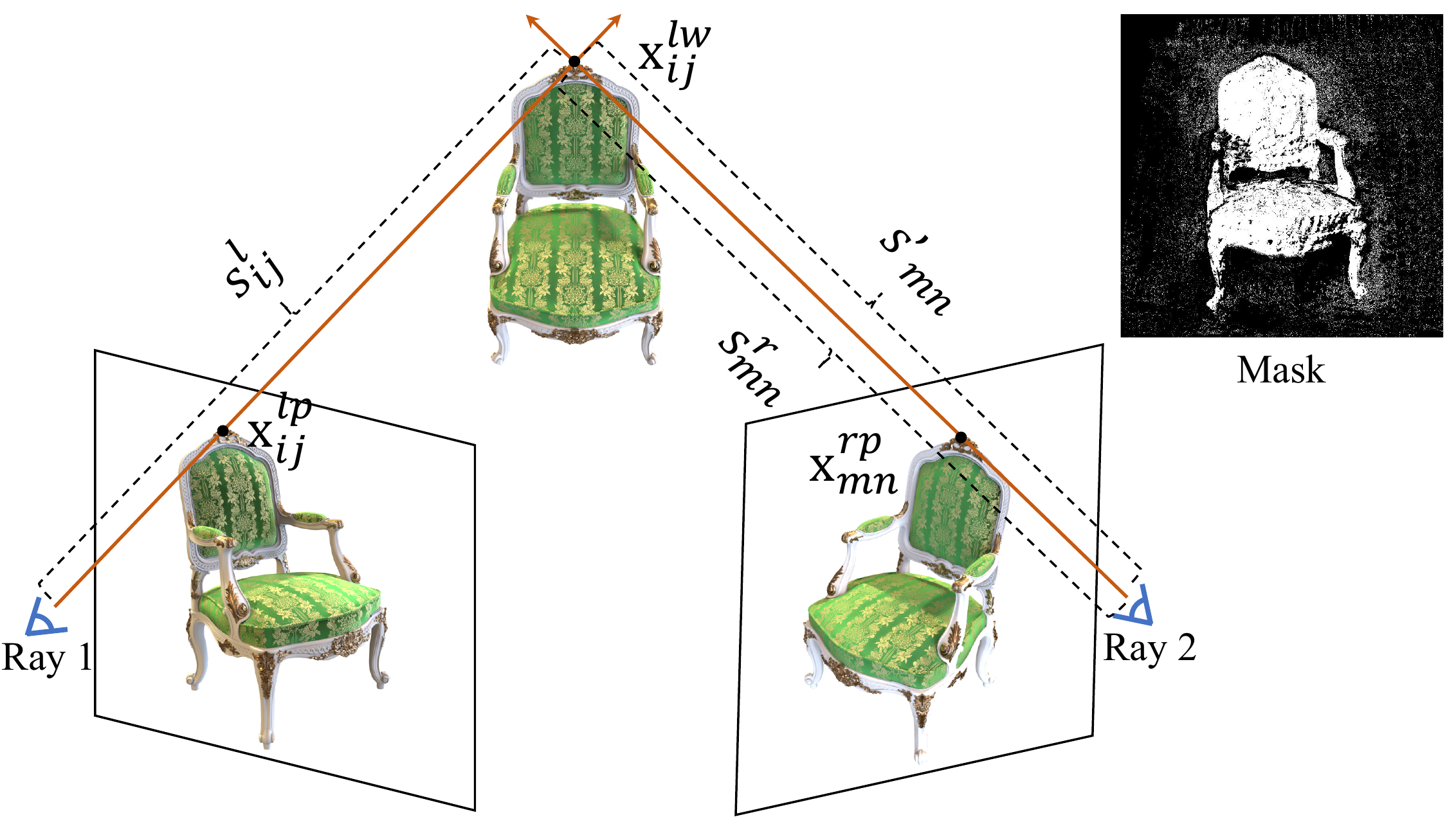}
    \caption{Illustration of deriving Depth-based Mask. We first derive the world coordinate $\mathbf{x}_{ij}^{lw}$ of pixel $(i,j)$ in the left view and then project the world coordinate $\mathbf{x}_{ij}^{lw}$ into the right view, which leads to pixel $(m,n)$. If the difference of projected depth $s_{mn}^{'}$ and depth label $s_{mn}^{r}$ is less than a threshold $\alpha$, the pixel $(i,j)$ and $(m,n)$ are marked as pixels satisfying 3D correspondence. By convention, depth is defined as the coordinate value along the z-axis in the corresponding camera coordinate system.}
\label{fig:mask_c2w}
\end{figure*}

\begin{definition}[Multi-view Appearance Consistency]
\label{def: appearance}
The multi-view appearance consistency refers to the color difference between the pixel $(i,j)\in\mathcal{M}$ (in the left view of Fig.~\ref{fig:mask_c2w}) and its corresponding pixel $(m,n):=\mathcal{T}((i,j))$ (in the right view of Fig.~\ref{fig:mask_c2w}) should be smaller than a threshold value $\epsilon_c$, i.e.:
\begin{equation}
\begin{aligned}
||C_{\theta}(\mathbf{r}_{ij}) - C_{\theta}(\mathbf{r}_{mn})||_{2}^{2} \leq \epsilon_c,
\end{aligned}
\end{equation}
where $C_{\theta}(\mathbf{r}_{ij})$ and $C_{\theta}(\mathbf{r}_{mn})$ are color labels of pixel $(i,j)$ and $(m,n)$.
\end{definition} 

\subsection{Neural Radiance Fields with Multi-view 3D Consistency}
\label{subsec: multi-view 3D concsitency nerf}
Based on the importance of 3D correspondence, we propose ConsistentNeRF to enforce NeRF-based algorithms to focus on the 3D correspondence relationship. 
Given a series of images for the specific scenario, it masks pixels satisfying 3D correspondence relationship between source views and target views.
With no loss of generality, we show the derivation of mask in two views.
As shown in Fig.~\ref{fig:mask_c2w}, ConsistentNeRF samples a bunch of pixels $\{(i,j)\}$ with coordinates $\{\mathbf{x}_{ij}^{lp}=[i, j, 1]^{T}\}$ in the left camera coordinate, where $l$ denotes the left camera view and $p$ denotes the pixel coordinate.
For each pixel $(i,j)$, one camera ray is cast from the camera origin $\mathbf{o}$ along with the ray direction $\mathbf{d}$. With the estimated depth $s_{ij}^l$ of pixel $(i,j)$ in the left camera view, the world coordinate of the intersection point $\mathbf{x}_{ij}^{lw}$ can be derived as
\begin{equation}
\label{eq:mask_1c2w}
\begin{aligned}
\mathbf{x}_{ij}^{lw} = (\mathbf{R}^{l})^{-1} \mathbf{K}^{-1} \cdot (s_{ij}^l \cdot \mathbf{x}_{ij}^{lp}),
\end{aligned}
\end{equation}
where $\mathbf{R}^{l}$ is the world-to-camera transformation matrix of the left camera view, $\mathbf{K}$ is the camera intrinsic matrix.

To get the pixel coordinate of the intersection point $\mathbf{x}_{ij}^{lw}$ in the right view, the estimated world coordinate $\mathbf{x}_{ij}^{lw}$ is transformed into the image plane of the right camera view with the world-to-camera transformation matrix $\mathbf{R}^{r}$ and camera intrinsic matrix $\mathbf{K}$ as follows:
\begin{equation}
\label{eq:mask_w2c2}
\begin{aligned}
s_{mn}^{'} \cdot \mathbf{x}_{mn}^{rc} = \mathbf{K} \mathbf{R}^{r} \mathbf{x}_{ij}^{lw},
\end{aligned}
\end{equation}
where $\mathbf{x}_{mn}^{rc} = (m, n, 1)$ is the pixel coordinate by projecting the intersection point $\mathbf{x}_{ij}^{lw}$ onto the right camera image plane,
$s_{mn}^{'}$ is the estimated depth of the intersection point $\mathbf{x}_{ij}^{lw}$ in the right camera.

Pixels $(i,j)$ and $(m,n)$ are masked as pixels with 3D correspondence relationship when 1) the pixel $(m,n)$ is not out of the boundary of the right image plane and 2) the transformed depth $s_{mn}^{'}$ and depth $s_{mn}^{r}$ of pixel $(m,n)$ are sufficiently close.
Pixel $(i,j)$ is regarded as a pixel that does not satisfy 3D correspondence under the sparse view setting and is excluded when it cannot find a pixel that satisfies the above condition in all training views. 
Following the above derivation, we set a threshold $\alpha$ to mask pixels with 3D correspondence relationship as follows:
\begin{equation}
\label{eq:mask_threshold}
|s_{mn}^{r} - s_{mn}^{'}| < \alpha \to \text{pixel $(i,j)$} \in \mathcal{M}, \mathcal{T}((i,j)) = ((m,n)),
\end{equation}
where $\mathcal{M}$ is defined to be the set containing masked pixels and $\mathcal{T}$ defines the correspondence relationship between pixel $(i,j)$ and $(m,n):=\mathcal{T}((i,j))$.

\begin{figure*}[t]
    \centering
    \includegraphics[width=18cm]{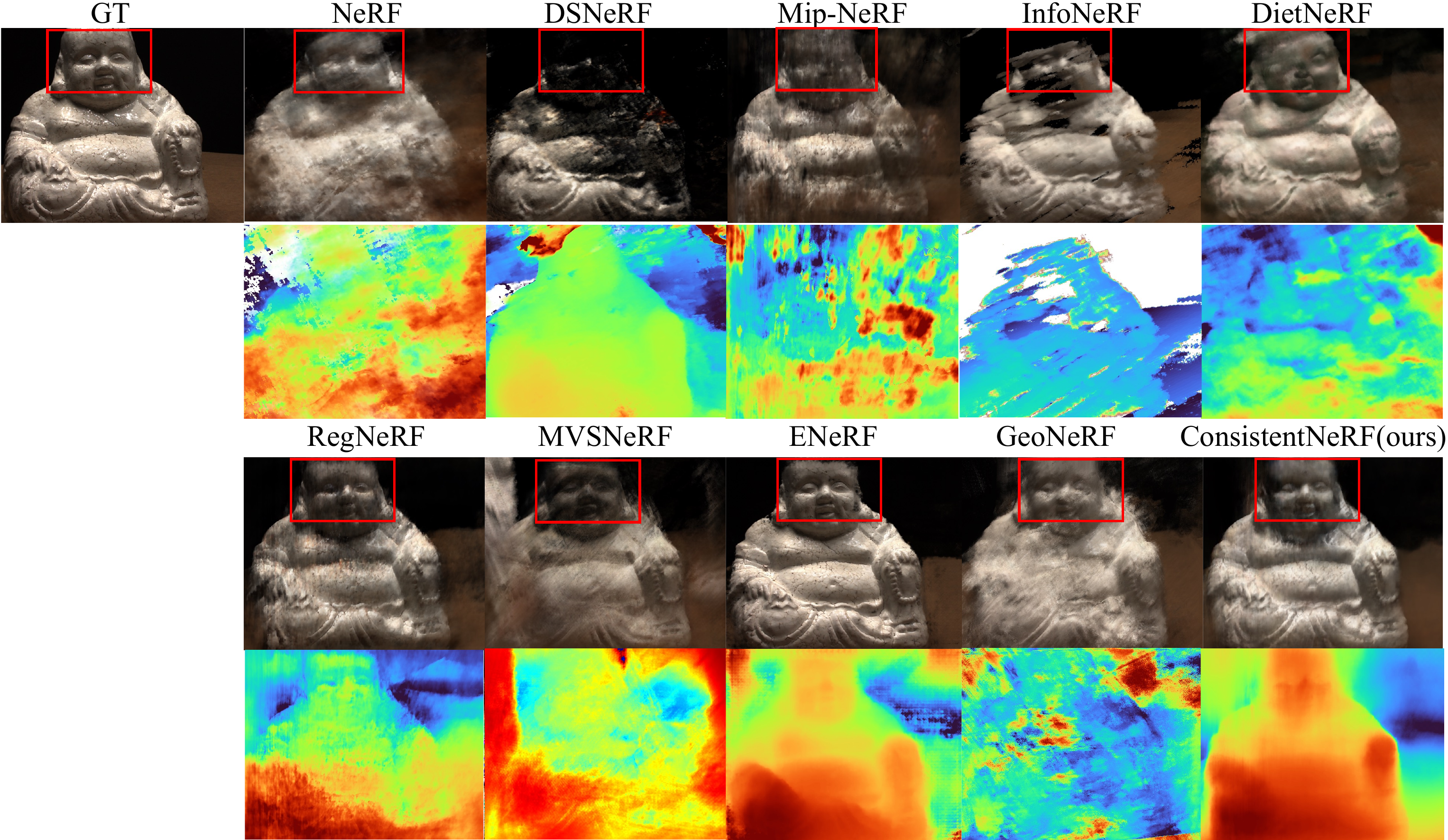}
    \caption{Novel View Synthesis Results on DTU data set with 3 views as input. We observe that the baselines suffer from blur results, while our ConsistentNeRF can produce sharp results with fine-grained details. } 
\label{fig:ConsistentNeRF_3view_dtu_fig}
\end{figure*}

\begin{figure*}[h]
    \centering
    \includegraphics[width=15cm]{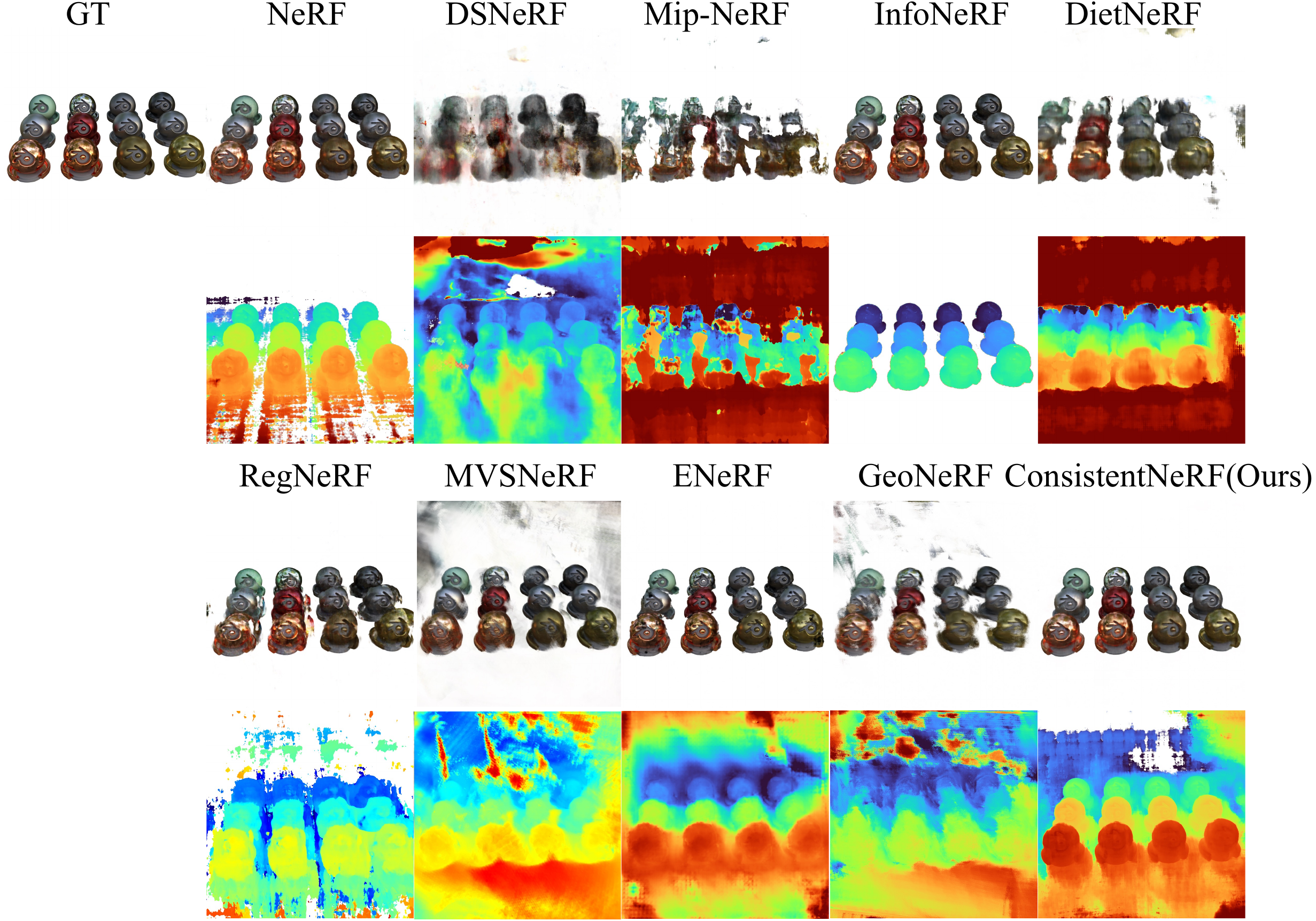}
    \caption{Novel View Synthesis Results on NeRF Synthetic data set with 3 views as input. We observe that the baselines suffer from blur results, while our ConsistentNeRF can produce sharp results with fine-grained details. } 
\label{fig:ConsistentNeRF_3view_nerf_fig}
\end{figure*}

\begin{figure*}[h]
    \centering
    \includegraphics[width=18cm]{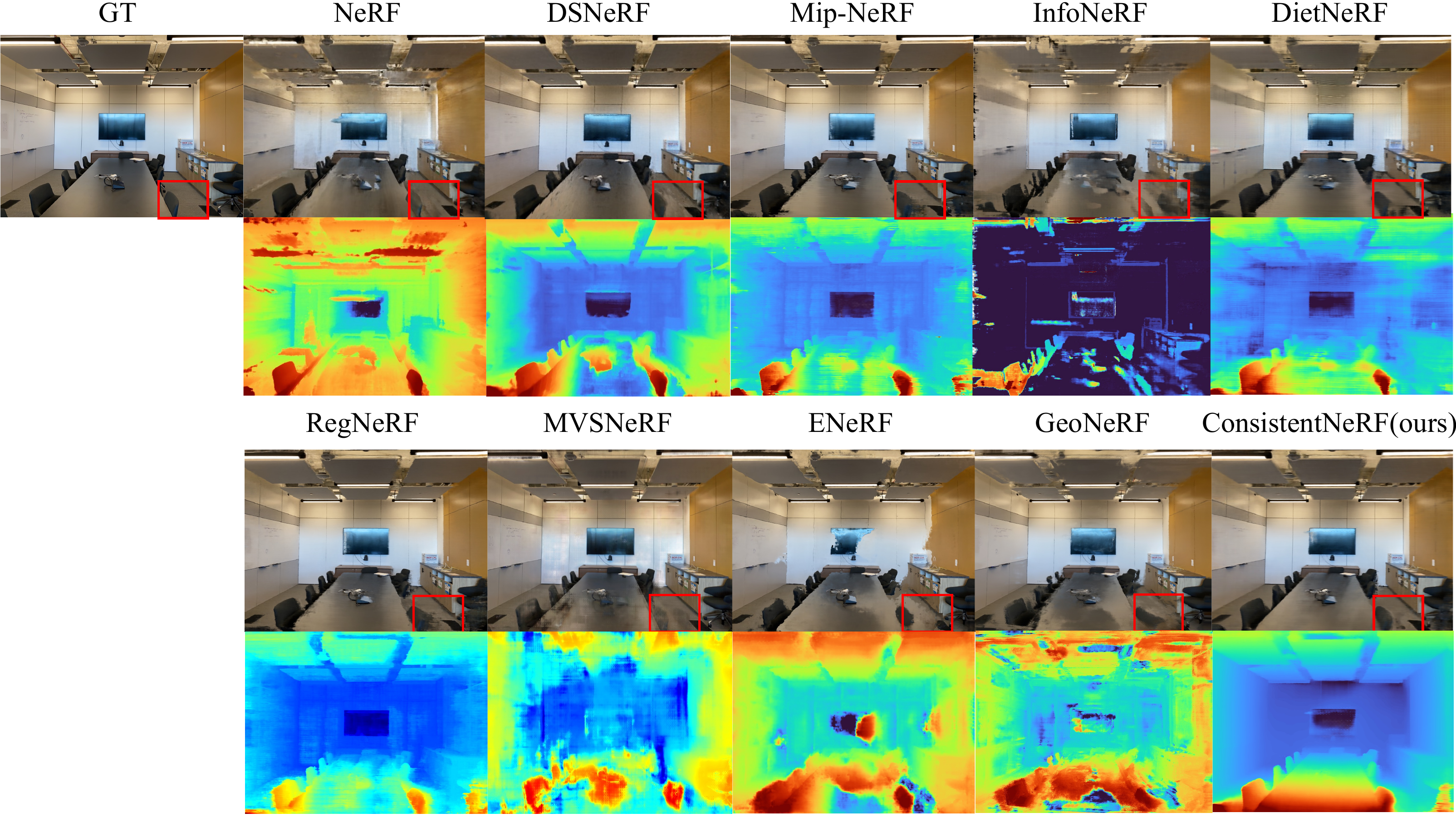}
    \caption{Novel View Synthesis Results on LLFF data set with 3 views as input. We observe that the baselines suffer from blur results, while our ConsistentNeRF can produce sharp results with fine-grained details.} 
\label{fig:ConsistentNeRF_3view_llff_fig}
\end{figure*}

With the derived mask, the loss function is defined as 
\begin{equation}
\label{eq:nerf_hardmask_l2_loss}
\begin{aligned}
\mathcal{L} = \frac{1}{|\mathcal{R}|}\sum_{\mathbf{r}\in \mathcal{R}\cap\mathcal{M}} ||\hat{C}_{\theta}(\mathbf{r}) - C(\mathbf{r})||_2^2 \\
+ \lambda\sum_{\mathbf{r}\notin \mathcal{R}\cap\mathcal{M}} ||\hat{C}_{\theta}(\mathbf{r}) - C(\mathbf{r})||_2^2,
\end{aligned}
\end{equation}
where $\mathcal{R}$ denotes the set of rays, the coefficient $\lambda \ll 1$ controls the loss ratio of emphasizing the pixels satisfying the correspondence relationship.

Note that according to Definition~\ref{def: appearance}, the predicted color difference between pixels $(i,j)$ and $(m,n)$ (in the left/right view of Fig.~\ref{fig:mask_c2w}) should be smaller than a threshold value $\epsilon_c$, i.e.,
\begin{equation}
\label{eq:Appearance_Regularization}
||\hat{C}_{\theta}(\mathbf{r}_{ij}) - \hat{C}_{\theta}(\mathbf{r}_{mn})||_{2}^{2} \leq \epsilon_c,
\end{equation}
where $\hat{C}_{\theta}(\mathbf{r}_{ij})$ and $\hat{C}_{\theta}(\mathbf{r}_{mn})$ are predicted colors of pixel $(i,j)$ and $(m,n) \in \mathcal{M}$.
As shown in Proposition~\ref{pro: appearance}, the above loss function, which focuses on the pixels selected by the mask, implicitly emphasizes the appearance consistency in the optimization of NeRF. The proof is provided in Appendix~\ref{app: appearance_consistency_pf}. We also show that it emphasizes the geometry consistency in the optimization of NeRF (see Appendix~\ref{app: geometry_consistency_pf} for more details).

\begin{proposition}[Multi-view Appearance Consistency]
\label{pro: appearance}
Directly minimizing the above appearance consistency leads to trivial solution $\hat{C}_{\theta}(\mathbf{r}_{ij})= \hat{C}_{\theta}(\mathbf{r}_{mn}) = 0$. 
Focusing on minimizing the errors between predicted color values and their ground truth for pixels included by the mask as in Eqn.~(\ref{eq:nerf_hardmask_l2_loss}) would help to emphasize the appearance consistency:
\begin{equation}
\label{eq:Appearance_Regularization_upper_bound}
\begin{aligned}
||\hat{C}_{\theta}(\mathbf{r}_{ij}) - C(\mathbf{r}_{ij})||_2^2  + ||\hat{C}_{\theta}(\mathbf{r}_{mn}) - C(\mathbf{r}_{mn})||_2^2 \\
\geq  \frac{1}{4}||\hat{C}_{\theta}(\mathbf{r}_{ij}) - \hat{C}_{\theta}(\mathbf{r}_{mn})||_{2}^{2} - \epsilon_c/2 .
\end{aligned}
\end{equation}
\end{proposition} 

The above estimated mask locates pixels satisfying 3D correspondence relationship, which enforces NeRF to focus on the optimization of 3D consistency. 

\subsection{Neural Radiance Fields with Single-view 3D consistency}
In addition to regularizing appearance and geometry consistency among different views, we also propose to regularize 3D consistency in the same view by using the depth predicted from state-of-the-art monocular depth estimation method MiDaS~\cite{ranftl2020towards} as additional supervision. 
Considering that the depth predicted from monocular depth estimation methods can not guarantee the scale-invariant property, we apply the depth-invariant geometry consistency regularization to measure the depth relationships between pixels in the same patch.
For predicted depth maps $s_{mn}^{'}$ and MiDas depths $s_{mn}^{r}$ of pixels in a patch, each with $N$ pixels indexed by $(m, n)$, we apply the scale-invariant mean squared depth error (in log space) defined in ~\cite{eigen2014depth}, i.e.,
\begin{equation}
\label{eq:Midas_depth_loss}
\begin{aligned}
D(s_{mn}^{'}, s_{mn}^{r}) &= \frac{1}{2N} \sum_{m,n}(\log{s_{mn}^{'}} - \log{s_{mn}^{r}} + \\
& \frac{1}{N}\sum_{m,n}(\log{s_{mn}^{r}} - \log{s_{mn}^{'}}))^2.
\end{aligned}
\end{equation}
For any depth prediction $s_{mn}^{'}$, $e^{\frac{1}{N}\sum_{m,n}(\log{s_{mn}^{r}} - \log{s_{mn}^{'}}))^2}$ is the scale that best aligns it to MiDas depth.
Our intuition behind the above idea is that although the scale of the predicted depth from MiDas models is not accurate, the local structure (relative relationship) of predicted depth in a small patch contains a relatively accurate 3D consistency relationship, which can be used to regularize NeRF's optimization. 

\begin{table*}[t]
\caption{Performance (PSNR, SSIM and LPIPS) comparison among state-of-the-art NeRF methods on DTU, NeRF Synthetic and Forward-Facing data sets. $\uparrow$ means the larger is better; $\downarrow$ means the smaller is better.}
\centering
\label{tab: ConsistentNeRF_redult}
\setlength{\tabcolsep}{1.0mm}{
\begin{tabular}{c|c|c|ccc|ccc|ccc}
\hline\hline
\multirow{2}*{Method} & \multirow{2}*{Setting} & \multirow{2}*{Pretrain} & \multicolumn{3}{c|}{Real Data (DTU)} & \multicolumn{3}{c|}{Synthetic Data (NeRF)} & \multicolumn{3}{c}{Forward-Facing (LLFF)}\\
\cline{4-12}
~ & ~ & ~ & PSNR$\uparrow$ & SSIM$\uparrow$ & LPIPS$\downarrow$ & PSNR$\uparrow$ & SSIM$\uparrow$ & LPIPS$\downarrow$ & PSNR$\uparrow$ & SSIM$\uparrow$ & LPIPS$\downarrow$ \\
\hline\hline
NeRF~\cite{mildenhall2020nerf} & \multirow{10}*{\tabincell{c}{3-view}} & \xmark & 11.40 & 0.50 & 0.49 & 14.59 & 0.82 & 0.29 & 12.52 & 0.34 & 0.60 \\
DSNeRF~\cite{deng2021depth} & ~ & \xmark & 11.80 & 0.52 & 0.49 & 15.13 & 0.82 & 0.30 & 13.10 & 0.35 & 0.62 \\
Mip-NeRF~\cite{barron2021mip} & ~ & \xmark & 15.87 & 0.73 & 0.42 & 16.52 & 0.80 & 0.28 & 20.19 & 0.71 & 0.47 \\
InfoNeRF~\cite{kim2022infonerf} & ~ & \xmark & 17.54 & 0.62 & 0.44 & 14.51 & 0.75 & 0.30 & 16.78 & 0.47 & 0.56 \\ 
DietNeRF~\cite{jain2021putting} & ~ & \xmark & 12.94 & 0.42 & 0.64 & 17.55 & 0.77 & 0.28 & 19.84 & 0.58 & 0.51 \\
RegNeRF~\cite{niemeyer2021regnerf} & ~ & \xmark & 21.57 & 0.84 & \textbf{0.31} & 17.39 & 0.82 & 0.26 & 20.36 & 0.72 & 0.45 \\
MVSNeRF~\cite{chen2021mvsnerf} & ~ & \cmark & 19.17 & 0.80 & 0.34 & 15.12 & 0.82 & 0.29 & 18.99 & 0.68 & \textbf{0.41} \\
GeoNeRF~\cite{johari2022geonerf} & ~ & \cmark & 16.51 & 0.56 & 0.43 & 17.67 & 0.73 & 0.33 & 17.76 & 0.50 & 0.49 \\
ENeRF~\cite{lin2022efficient} & ~ & \cmark & 18.65 & 0.83 & 0.40 & 18.14 & \textbf{0.83} & \textbf{0.20} & 20.30 & \textbf{0.75} & 0.45 \\
ConsistentNeRF (\textbf{Ours}) & ~ & \xmark & \textbf{22.14} & \textbf{0.88} & 0.34 & \textbf{19.63} & \textbf{0.83} & \textbf{0.20} & \textbf{21.77} & 0.73 & 0.43 \\
\hline\hline
\end{tabular}}
\end{table*}

\section{Experiments}
\label{sec: exp}
\textbf{Datasets.} 

We evaluate the proposed method on three diverse datasets, namely the real-world multi-view DTU dataset~\cite{jensen2014large}, Forward-Facing LLFF dataset~\cite{mildenhall2019local} and Realistic Synthetic NeRF dataset~\cite{mildenhall2020nerf}. Specifically, we follow PixelNeRF~\cite{yu2021pixelnerf} to split the DTU dataset into 88 training scenes and 16 testing scenes. We utilize the 88 training scenes to pre-train the IBRNet~\cite{wang2021ibrnet} and MVSNeRF~\cite{chen2021mvsnerf} models. For each testing scene across the three datasets, we follow MVSNeRF to select three views from 20 nearby views as training views, and four views as testing views. In accordance with prior NeRF techniques, we evaluate all the methods on the DTU dataset with object masks applied to the rendered and ground truth images.

\textbf{Evaluation Metrics.} For performance comparison, we report the mean of peak signal-to-noise ratio (PSNR) ~\cite{sara2019image}, structural similarity index (SSIM)~\cite{wang2004image} and Learned Perceptual Image Patch Similarity (LPIPS) perceptual metric~\cite{zhang2018unreasonable}.

\textbf{Implementation Details.} 
We compare our method with NeRF based methods, including NeRF~\cite{mildenhall2020nerf}, DSNeRF~\cite{deng2021depth}, Mip-NeRF~\cite{barron2021mip}, InfoNeRF~\cite{kim2022infonerf}, DietNeRF~\cite{jain2021putting}, RegNeRF~\cite{niemeyer2021regnerf}, MVSNeRF~\cite{chen2021mvsnerf}, GeoNeRF~\cite{johari2022geonerf}, and ENeRF~\cite{lin2022efficient}. For all NeRF~\cite{mildenhall2020nerf} based methods which do not require pre-training, we directly train the model from scratch for each target scene. In our experiments, we use the depth extracted from a pre-trained MVSNeRF~\cite{chen2021mvsnerf} to derive the mask. The depth also serves as the supervision for DSNeRF~\cite{deng2021depth} and our methods for fair comparisons. For MiDas Depth, we use the DPT Large pre-trained model to derive the monocular depth information~\cite{ranftl2020towards}. All mentioned methods (NeRF, DSNeRF, ConsistentNeRF) are trained with 50,000 iterations. For multi-view 3D consistency constraint, the threshold $\alpha$ is set to be 0.1 and $\lambda$ is set to be 0.1 on DTU, LLFF and NeRF Synthetic data set. We run each method with four random seeds and report the mean results. More implementation details are provided in Appendix~\ref{app:details}.

\begin{table*}[t]
\caption{Ablation study on ablating two consistency regularizations on the LLFF data set with 3 training views as input. For performance (PSNR, SSIM and LPIPS) comparison, $\uparrow$ means the larger is better; $\downarrow$ means the smaller is better.}
\centering
\label{tab: ablation_mask}
\begin{tabular}{c|ccc}
\hline\hline
\multirow{2}*{Method} & \multicolumn{3}{c}{Forward-Facing (LLFF)} \\
\cline{2-4}
& PSNR$\uparrow$ & SSIM$\uparrow$ & LPIPS$\downarrow$ \\
\hline\hline
ConsistentNeRF & 21.77 & 0.73 & 0.43 \\
w/o Single-view Consistency & 20.75 & 0.73 & 0.44 \\
w/o Multi-view Consistency & 20.85 & 0.73 & 0.44 \\
w/o Single-view and Multi-view Consistency & 20.36 & 0.72 & 0.45 \\
\hline\hline
\end{tabular}
\end{table*}

\textbf{Initialization for Stable Optimization.} 
During our experiments, we observe that NeRF is prone to a catastrophic failure at the initialization stage in which MLP emits negative values before the ReLU activation. In this case, all predicted $\sigma$ values are zero, and gradients back-propagated from the loss function to MLP parameters are zero, leading to the failure of the optimization. To address the above failure, Mip-NeRF~\cite{barron2021mip} proposes to use the softplus function to stabilize the optimization. However, we observe that NeRF overfits to training views by using the softplus function in the sparse view setting. In this paper, we propose to modify the initialization of bias parameters in the MLP to guarantee both stable optimization and good generalization ability. During our experiments, we find that initializing the value of bias parameters in MLP using a uniform distribution between 0 and 1 leads to acceptable results. The comparison results are reported in Appendix~\ref{app: degenerate of NeRF}.

\subsection{View Synthesis Results}

In the experiment, we evaluate the performance achieved by the above-mentioned NeRF models under sparse view settings and compare them with our proposed ConsistentNeRF. Quantitative results are shown in Tab.~\ref{tab: ConsistentNeRF_redult}. 
For 3 input view settings, our proposed ConsistentNeRF could largely improve the performance of the original NeRF, e.g., 70\% relative PSNR improvement is achieved on the DTU data set. Besides, when compared with DSNeRF which directly introduces depth constrain, our ConsistentNeRF could bring larger performance improvement through regularizing the optimization with 3D consistency relationship.
When further compared with NeRF-based methods with additional regularization, like Mip-NeRF~\cite{barron2021mip}, InfoNeRF~\cite{kim2022infonerf}, DietNeRF~\cite{jain2021putting}, RegNeRF~\cite{niemeyer2021regnerf}, ConsistentNeRF consistently shows better performance.
Quantitative results in Fig.~\ref{fig:ConsistentNeRF_3view_dtu_fig}, Fig.~\ref{fig:ConsistentNeRF_3view_nerf_fig} and Fig.~\ref{fig:ConsistentNeRF_3view_llff_fig} also support the above claim.  

We also compare ConsistentNeRF with MVSNeRF~\cite{chen2021mvsnerf}, GeoNeRF~\cite{johari2022geonerf} and ENeRF~\cite{lin2022efficient}, which require the pre-training and per-scene optimization.
As shown in Tab.~\ref{tab: ConsistentNeRF_redult}, MVSNeRF, GeoNeRF and ENeRF produce better results than the vanilla NeRF in the 3 view setting. However, we still observe some inconsistent results when the testing view is far from the training views. 
For example, as shown in Fig.~\ref{fig:ConsistentNeRF_3view_dtu_fig}, Fig.~\ref{fig:ConsistentNeRF_3view_nerf_fig} and Fig.~\ref{fig:ConsistentNeRF_3view_llff_fig}, these methods produce images with blur results and poor lighting, while our proposed ConsistentNeRF can predict more sharp results and the correct lighting effect of the pixels in the target view using the multi-view and single-view 3D consistency constraint.

\subsection{Ablation Study}
As shown in Tab.~\ref{tab: ablation_mask}, we ablate the performance of multi-view and single-view 3D consistency regularization introduced in Sec.~\ref{subsec: multi-view 3D concsitency nerf}. 
With either multi-view 3D consistency regularization or single-view 3D consistency regularization, ConsistentNeRF consistently outperforms the baseline model in all metrics.
Adding both multi-view and single-view 3D consistency regularization leads to the best performance.

\section{Limitation}
One limitation of our paper is that we adopt a pre-trained MVSNeRF to derive the mask information for multi-view 3D consistency regularization.
In real-world applications, it is hard to derive the mask information from a pre-trained NeRF model like MVSNeRF. 
In the future work, one potential direction is to apply flow model~\cite{teed2020raft} which directly utilizes RGB information to derive the correspondence relationship among pixels in different views. 
The other potential direction is to extend our framework to RGBD settings to derive the correspondence relationship among pixels in different views using the depth information from RGBD tensors. 
In addition, similar to most NeRF-based methods, our proposed optimization can not render images with high quality when the target view is far from source views as 3D correspondence relationship is hard to utilize in this case.

\section{Conclusion}
In this paper, we target to the challenging sparse view synthesis problem and proposed ConsistentNeRF, which enhances Neural Radiance Fields with 3D Consistency.
To build correspondences among pixels in different views, we propose a mask-based loss that locates the pixels with 3D consistency, instead of treating all pixels equally in the training objective. Moreover, we adopt a depth consistency regularization among pixels in the same patch to regularize the 3D consistency among pixels in the same view. Our experimental results demonstrate that our proposed methods significantly improve the performance of representative NeRF methods with sparse view settings and can bring larger performance improvement than previous depth-based methods. These promising results suggest that consistency-based NeRF is an important direction for rendering images with both correct geometry and fine-grained details. 
In conclusion, our proposed methods offer a new and effective solution to the challenging problem of sparse view synthesis and have promising potential for future applications in various fields.

\newpage

{\small
\bibliographystyle{ieee_fullname}
\bibliography{egbib}
}


\newpage

\appendix
\section*{Appendix}

\section{Multi-view 3D Appearance Consistency}
\label{app: appearance_consistency_pf}

\begin{definition}[Multi-view Appearance Consistency]
The multi-view appearance consistency refers to the color difference between the pixel $(i,j)\in\mathcal{M}$ (in the left view of Fig.~\ref{fig:mask_c2w}) and its corresponding pixel $(m,n):=\mathcal{T}((i,j))$ (in the right view of Fig.~\ref{fig:mask_c2w}) should be smaller than a threshold value $\epsilon_c$, i.e.:
\begin{equation}
\begin{aligned}
||C_{\theta}(\mathbf{r}_{ij}) - C_{\theta}(\mathbf{r}_{mn})||_{2}^{2} \leq \epsilon_c,
\end{aligned}
\end{equation}
where $C_{\theta}(\mathbf{r}_{ij})$ and $C_{\theta}(\mathbf{r}_{mn})$ are color labels of pixel $(i,j)$ and $(m,n)$.
\end{definition} 

\begin{definition}[Consistency of Estimated Appearance]
\label{def: appearance_est2}
The multi-view 3D consistency of estimated appearance refers to 
the predicted color difference between pixel $(i,j)\in\mathcal{M}$ and pixel $(m,n):=\mathcal{T}((i,j))$ (in the left/right view of Fig.~\ref{fig:mask_c2w}) should be smaller than a 
threshold value $\epsilon_c$, i.e.:
\begin{equation}
\label{eq:app_Appearance_Regularization}
\begin{aligned}
||\hat{C}_{\theta}(\mathbf{r}_{ij}) - \hat{C}_{\theta}(\mathbf{r}_{mn})||_{2}^{2} \leq \epsilon_c,
\end{aligned}
\end{equation}
where $\hat{C}_{\theta}(\mathbf{r}_{ij})$ and $\hat{C}_{\theta}(\mathbf{r}_{mn})$ are predicted color of pixel $(i,j)$ and $(m,n)$.
\end{definition}

\begin{proposition}[Multi-view Appearance Consistency]
Directly minimizing appearance consistency in Definition \ref{def: appearance_est2} leads to trivial solution $\hat{C}_{\theta}(\mathbf{r}_{ij})= \hat{C}_{\theta}(\mathbf{r}_{mn}) = 0$. 
Focusing on minimizing the errors between predicted color values and their ground truth for pixels included by the Hard-Mask as in Eqn.~(\ref{eq:nerf_hardmask_l2_loss}) would help to emphasize the appearance consistency:

\begin{align*}
& ||\hat{C}_{\theta}(\mathbf{r}_{ij}) - C(\mathbf{r}_{ij})||_2^2  + ||\hat{C}_{\theta}(\mathbf{r}_{mn}) - C(\mathbf{r}_{mn})||_2^2 \\
&\geq \frac{1}{4}||\hat{C}_{\theta}(\mathbf{r}_{ij}) - \hat{C}_{\theta}(\mathbf{r}_{mn})||_{2}^{2} - \epsilon_c/2 .
\end{align*}
\end{proposition} 

Proof: 
\begin{align*}
    &2||\hat{C}_{\theta}(\mathbf{r}_{ij}) - C(\mathbf{r}_{ij})||_2^2  + 2||\hat{C}_{\theta}(\mathbf{r}_{mn}) - C(\mathbf{r}_{mn})||_2^2\\
    \ge& \|(\hat{C}_{\theta}(\mathbf{r}_{ij}) - \hat{C}_{\theta}(\mathbf{r}_{mn})) + (C(\mathbf{r}_{mn})- C(\mathbf{r}_{ij})) \|_2^2\\
    \ge& \frac{1}{2}\|\hat{C}_{\theta}(\mathbf{r}_{ij}) - \hat{C}_{\theta}(\mathbf{r}_{mn})\|_2^2 - \|C(\mathbf{r}_{mn})- C(\mathbf{r}_{ij})\|_2^2
\end{align*}
The first inequality follows from the fact that for two vectors $\mathbf{a},\mathbf{b}$,
\begin{align*}
    2\|\mathbf{a}\|_2^2 + 2\|\mathbf{b}\|_2^2 - \|\mathbf{a}+\mathbf{b}\|_2^2=\|\mathbf{a}-\mathbf{b}\|_2^2 \ge 0.
\end{align*}
The second inequality is due to the fact that
\begin{align*}
    2\|\mathbf{a}+\mathbf{b}\|_2^2 - (\|\mathbf{a}\|_2^2-2\|\mathbf{b}\|_2^2)=\|\mathbf{a}+2\mathbf{b}\|_2^2 \ge 0.
\end{align*}

\section{Multi-view 3D Geometry Consistency}
\label{app: geometry_consistency_pf}

\begin{definition}[Multi-view Geometry Consistency]
The geometry consistency refers to the depth difference between the depth of pixel $(m, n)\in\mathcal{M}$ in the right camera view and the depth generated by warping its corresponding pixel $(i,j):=\mathcal{T}((m,n))$ from left camera to the right camera should be smaller than a threshold value $\epsilon_s$, i.e.:
\begin{equation}
\label{eq:def_Geometry_Consistency}
\begin{aligned}
||s_{mn}^{r} - s_{mn}^{'}||_{2}^{2} \leq \epsilon_s,
\end{aligned}
\end{equation}
where $s_{mn}^{r}$ is the depth for pixel $(m, n)$ and $s_{mn}^{'}$ is the projected depth from left camera pixel $(i, j)$.
\end{definition} 

\begin{definition}[Consistency of Estimated Geometry]
The consistency of estimated geometry refers to the predicted depth difference between the depth of pixel $(m, n)\in\mathcal{M}$ in the right camera view and the predicted depth generated by warping its corresponding pixel $(i,j):=\mathcal{T}((m,n))$ from left camera to the right camera should be smaller than a threshold value $\epsilon_s$, i.e.:
\begin{equation}
\label{eq:Geometry_Regularization}
\begin{aligned}
||\hat{s}_{\theta}(\mathbf{r}_{mn}) - \hat{s}_{\theta}^{'}(\mathbf{r}_{mn})||_{2}^{2} \leq \epsilon_s,
\end{aligned}
\end{equation}
where $\hat{s}_{\theta}(\mathbf{r}_{mn})$ is the predicted depth for pixel $(m, n)$ and $\hat{s}_{\theta}^{'}(\mathbf{r}_{mn})$ is the projected depth from left camera pixel $(i, j)$.
\end{definition}

\begin{proposition}[Multi-view Geometry Consistency] \label{pro: geometry}
Similar to multi-view Appearance Consistency Regularization, focusing on optimizing the error between predicted depth value and its ground truth for pixels included by Hard-Mask as in Eqn. (\ref{eq:nerf_hardmask_l2_loss}) would help to emphasize the geometry consistency:
\begin{equation}
\label{eq:Geometry_Regularization_upper_bound}
\begin{aligned}
& ||\hat{s}_{\theta}(\mathbf{r}_{mn}) - s^{r}_{mn}||_2^2 +  ||\hat{s}_{\theta}^{'}(\mathbf{r}_{mn}) - s^{'}_{mn}||_2^2\\
&\geq \frac{1}{4}||\hat{s}_{\theta}(\mathbf{r}_{mn}) - \hat{s}_{\theta}^{'}(\mathbf{r}_{mn})||_{2}^{2} - \epsilon_s/2,
\end{aligned}
\end{equation}
\end{proposition}

Proof:

\begin{align*}
    &2||\hat{s}_{\theta}(\mathbf{r}_{mn}) - s^{r}_{mn}||_2^2  + 2||\hat{s}_{\theta}^{'}(\mathbf{r}_{mn}) - s^{'}_{mn}||_2^2\\
    \ge& \|(\hat{s}_{\theta}(\mathbf{r}_{mn}) - \hat{s}_{\theta}^{'}(\mathbf{r}_{mn})) + (s^{r}_{mn}- s^{'}_{mn}) \|_2^2\\
    \ge& \frac{1}{2}\|\hat{s}_{\theta}(\mathbf{r}_{mn}) - \hat{s}_{\theta}^{'}(\mathbf{r}_{mn})\|_2^2 - \|s^{r}_{mn}- s^{'}_{mn}\|_2^2
\end{align*}
The first inequality follows from the fact that for two vectors $\mathbf{a},\mathbf{b}$,
\begin{align*}
    2\|\mathbf{a}\|_2^2 + 2\|\mathbf{b}\|_2^2 - \|\mathbf{a}+\mathbf{b}\|_2^2=\|\mathbf{a}-\mathbf{b}\|_2^2 \ge 0.
\end{align*}
The second inequality is due to the fact that
\begin{align*}
    2\|\mathbf{a}+\mathbf{b}\|_2^2 - (\|\mathbf{a}\|_2^2-2\|\mathbf{b}\|_2^2)=\|\mathbf{a}+2\mathbf{b}\|_2^2 \ge 0.
\end{align*}

\section{Preliminary Study}
\label{app:preliminary}

By utilizing the homography warping relationship, we locate pixels satisfying 3D correspondence relationship.
Based on the masked pixels among training views, we find the respective 3D points and randomly sample different portions (30\%, 60\%, 100\%) of 3D points for the purpose of emphasizing the 3D correspondence.
We conduct each experiment using 4 random seeds and report the mean results.

\section{Implementation Details}
\label{app:details}

All our models are trained on the NVIDIA Tesla V100 Volta GPU cards. 
The NeRF-based models are implemented based on the code from~\cite{lin2020nerfpytorch}.
For MVSNet, we follow the released code and checkpoint to pre-train and finetune the models.
For Mask introduced in Sec.~\ref{subsec: multi-view 3D concsitency nerf}, we generate the mask information for each training image based on the correspondence among pixels in all training views.

\section{Solutions to Avoid Degenerate Results in NeRF}
\label{app: degenerate of NeRF}

As mentioned in Sec.~\ref{subsec: multi-view 3D concsitency nerf}, NeRF is prone to a catastrophic failure at the initialization stage in which MLP emits negative values before the ReLU activation. 
To address this issue, Mip-NeRF~\cite{barron2021mip} proposed to use the softplus function to yield a stable optimization process.
However, we observe that NeRF overfits training views by using the softplus function in the sparse view setting.
One possible reason could be that the predicted alpha value of sampled points should be sparse and dropping small values with ReLU activation could effectively improve the generalization ability. 
Based on the above consideration, we instead propose to modify the initialization of bias parameters in the MLP to guarantee both stable optimization and good generalization ability.
As shown in Tab.~\ref{tab: ablation_solution_avoid_degenerate_result}, our proposed initialization effectively improves the performance of NeRF and avoid the degenerate results when compared with SoftPlus activation and the original NeRF setting.

\begin{table}[h]
\setlength{\abovecaptionskip}{0cm}
\caption{Performance (PSNR, SSIM and LPIPS) comparison between SoftPlus and our proposed stable initialization to avoid degenerate results in NeRF on the DTU data set with 3 training views as input. $\uparrow$ means the larger is better; $\downarrow$ means the smaller is better.}
\centering
\label{tab: ablation_solution_avoid_degenerate_result}
\begin{tabular}{c|ccc}
\hline\hline
\multirow{2}*{Method} & \multicolumn{3}{c}{Real Data (DTU)} \\
\cline{2-4}
~ & PSNR$\uparrow$ & SSIM$\uparrow$ & LPIPS$\downarrow$  \\
\hline\hline
ReLU & 11.40 & 0.50 & 0.49 \\
SoftPlus & 14.26 & 0.68 & 0.45 \\
Stable Initialization & 16.91 & 0.73 & 0.41 \\
\hline\hline
\end{tabular}
\end{table}

\end{document}